# eRAKI: Fast Robust Artificial neural networks for K-space Interpolation (RAKI) with Coil Combination and Joint Reconstruction

Heng Yu[1], Zijing Dong[2,3], Yamin Arefeen[2], Congyu Liao[4], Kawin Setsompop[4,5], and Berkin Bilgic[3,6,7]

[1]Department of Automation, Tsinghua University, Beijing, China, [2]Department of Electrical Engineering and Computer Science, Massachusetts Institute of Technology, Cambridge, MA, United States, [3]Athinoula A. Martinos Center for Biomedical Imaging, Charlestown, MA, United States, [4]Radiological Sciences Laboratory, Stanford University, Stanford, CA, United States, [5]Department of Electrical Engineering, Stanford University, Stanford, CA, United States, [6]Harvard Medical School, Boston, MA, United States, [7]Harvard-MIT Health Sciences and Technology, Massachusetts Institute of Technology, Cambridge, MA, United States

## Synopsis

RAKI can perform database-free MRI reconstruction by training models using only auto-calibration signal (ACS) from each specific scan. As it trains a separate model for each individual coil, learning and inference with RAKI can be computationally prohibitive, particularly for large 3D datasets. In this abstract, we accelerate RAKI by more than 200 times by directly learning a coil-combined target and further improve the reconstruction performance using joint reconstruction across multiple echoes together with an elliptical-CAIPI sampling approach. We further deploy these improvements in quantitative imaging and rapidly obtain $T_2$ and $T_2^*$ parameter maps from a fast EPTI scan.

## Introduction

A number of machine learning approaches have been proposed for MRI reconstruction that allow high-quality reconstructions from highly undersampled data [1]. A promising approach in this direction is RAKI [2], which trains networks solely using autocalibration signal (ACS) data from each specific scan, without relying on a large training dataset. A drawback of RAKI is in the training and reconstruction time which can be prohibitively long as it needs to learn a model for real and imaginary components of each individual coil (e.g., 64 models for 32 channel data). This may be particularly problematic for large 3D datasets, impeding its clinical adoption. Self-consistent RAKI (sRAKI) [3,4] learns self-consistency using a single model for all the coils but the use of iterative optimization during reconstruction can still be time-consuming. In this abstract, we propose eRAKI to speed up RAKI by directly learning to reconstruct coil-combined target data using ESPIRiT-based sensitivity estimation[5]. We perform the learning/reconstruction in 3D k-space to benefit from 3D convolutions while utilizing the fully-sampled readout dimension to effectively increase the size of available ACS data. We show that the proposed eRAKI accelerates RAKI's computational speed by more than 200 times while improving reconstruction performance by taking the advantage of multi-echo joint reconstruction and elliptical-CAIPI [10] sampling. We further demonstrate the ability of eRAKI to speed-up the reconstruction of echo planar time-resolved imaging (EPTI)[9] acquisition by 10x times to provide distortion-free $T_2$ and $T_2^*$ parameter maps.

## Methods and Experiments

*eRAKI Reconstruction*

We use ESPIRiT to estimate coil sensitivities $C$ from ACS data $y_{acs}$ and perform coil combination to obtain single-channel ACS data in k-space, $y_{combo}$. We take $y_{combo}$ as the network's target to learn and train a single model using the following objective function:

$$argmin_\theta (\alpha ||y_{combo} - f_\theta(x_{acq})||_1 + (1-\alpha)||y_{combo} - f_\theta(x_{acq})||_2 + \beta ||\theta||_2)$$

where $f_\theta$ is the network, $\theta$ are the network parameters, $x_{acq}$ is the input under-sampled k-space. Regularization terms ($\alpha$ = 0.5 and $\beta$ = 0.15) are added to speed up the convergence.
After training the network using ACS data, we apply the trained model to perform inference on the entire undersampled k-space and the network output is the reconstruction result. The whole process takes place in 3D k-space to benefit from 3D convolution to help compensate for any information loss during the coil combination process. This also further reduces the overall reconstruction time, as only a single network needs to be trained across the whole 3D imaging volume, rather than one model per slice. Figure 1 shows the implementation of the proposed eRAKI network.

*eRAKI with joint reconstruction and elliptical-CAIPI sampling*

The proposed eRAKI is applied on 3D ME-MPRAGE data [6,7] and compared with 2D GRAPPA [8] and 2D RAKI as shown in Figures 2 and 3. To further improve reconstruction performance, we apply elliptical-CAIPI sampling while maintaining the same acceleration rate, and extend eRAKI to perform joint reconstruction across three echoes by concatenating echoes in the coil dimension.

*eRAKI with EPTI*

EPTI [9] can rapidly acquire distortion- and blurring-free multi-contrast data using an efficient spatiotemporal CAIPI sampling $k_y - t$ and B0-informed GRAPPA-like reconstruction. We apply RAKI and eRAKI on EPTI data for each acquired slice in $k_x - k_y - t$ space. All the experiments were done on a workstation with an Nvidia TITAN V and an AMD Ryzen Threadripper 3970X 32-Core Processor.
Code/data are available at https://anonymous.4open.science/r/77eb09e7-8a65-4a57-b559-741d6f8d4f37/.

## Results

Figure 2 shows that eRAKI achieves comparable performance as RAKI and GRAPPA, while improving the reconstruction speed >200-fold. Figure 3 shows that the reconstruction performance of eRAKI can be further improved by combining multi-echo joint reconstruction and elliptical-CAIPI sampling. Figure 4 shows that eRAKI can be applied to more complicated and advanced sampling methods where GRAPPA-like reconstruction can work, like EPTI, and achieve comparable performance while speeding up the computation by 10-fold. The table in Figure 5 summarizes the reconstruction times of the tested algorithms for these two datasets.

## Discussion and Conclusion

We accelerate RAKI reconstruction dramatically using a coil-combined target. For this, we capitalized on the fact that coil combination is a convolution in k-space, and can be well represented by a convolutional layer. Figure 5 shows the learning/inference time of each method. GRAPPA is implemented on CPU and RAKI and eRAKI run on GPU. Since GRAPPA needs to learn a model for each coil, it may not benefit significantly from a GPU implementation. In conclusion, eRAKI can reconstruct 3D volumes in a short timeframe and may facilitate the adoption of scan-specific deep learning on clinical scanners.

# Acknowledgements

This work was supported by research grants NIH R01 EB028797, U01 EB025162, P41 EB030006, U01 EB026996 and the NVidia Corporation for computing support.

# References

1. Knoll F, Hammernik K, Zhang C, et al. Deep-learning methods for parallel magnetic resonance imaging reconstruction: A survey of the current approaches, trends, and issues[J]. IEEE Signal Processing Magazine, 2020, 37(1): 128-140.

2. Akçakaya M, Moeller S, Weingärtner S, et al. Scan-specific robust artificial-neural-networks for k-space interpolation (RAKI) reconstruction: Database-free deep learning for fast imaging[J]. Magnetic resonance in medicine, 2019, 81(1): 439-453.

3. Hosseini S A H, Zhang C, Weingärtner S, et al. Accelerated coronary MRI with sRAKI: A database-free self-consistent neural network k-space reconstruction for arbitrary undersampling[J]. Plos one, 2020, 15(2): e0229418.

4. Hosseini S A H, Moeller S, Weingärtner S, et al. Accelerated coronary MRI using 3D SPIRiT-RAKI with sparsity regularization[C]//2019 IEEE 16th International Symposium on Biomedical Imaging (ISBI 2019). IEEE, 2019: 1692-1695.

5. Uecker M, Lai P, Murphy M J, et al. ESPIRiT—an eigenvalue approach to autocalibrating parallel MRI: where SENSE meets GRAPPA[J]. Magnetic resonance in medicine, 2014, 71(3): 990-1001.

6. van der Kouwe A J W, Benner T, Salat D H, et al. Brain morphometry with multiecho MPRAGE[J]. Neuroimage, 2008, 40(2): 559-569.

7. Mugler III J P, Brookeman J R. Three-dimensional magnetization-prepared rapid gradient-echo imaging (3D MP RAGE)[J]. Magnetic resonance in medicine, 1990, 15(1): 152-157.

8. Griswold M A, Jakob P M, Heidemann R M, et al. Generalized autocalibrating partially parallel acquisitions (GRAPPA)[J]. Magnetic Resonance in Medicine: An Official Journal of the International Society for Magnetic Resonance in Medicine, 2002, 47(6): 1202-1210.

9. Wang F, Dong Z, Reese T G, et al. Echo planar time-resolved imaging (EPTI)[J]. Magnetic resonance in medicine, 2019, 81(6): 3599-3615.

10. Breuer F A, Blaimer M, Heidemann R M, et al. Controlled aliasing in parallel imaging results in higher acceleration (CAIPIRINHA) for multi-slice imaging[J]. Magnetic Resonance in Medicine: An Official Journal of the International Society for Magnetic Resonance in Medicine, 2005, 53(3): 684-691.

# Figures

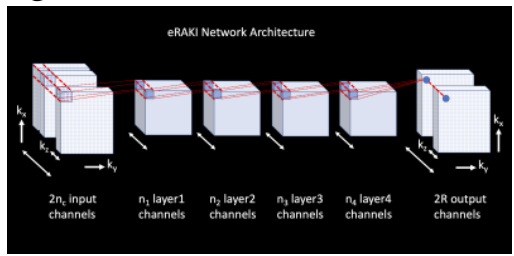

Our eRAKI uses 5-layer 3D CNN architecture. Since the learning target is coil-combined data, we need to learn R lines when the acceleration rate is R. The output channel is 2R so we only need to use one model to reconstruct all the data. Layer4 was followed by rectifier linear units (ReLU) as activation functions and we set $n_1=n_2=n_3=n_4=64$. The kernel sizes of the layers are 3×3×7, 1×1×5, 1×1×5, 1×1×3, and 1×1×1, respectively.

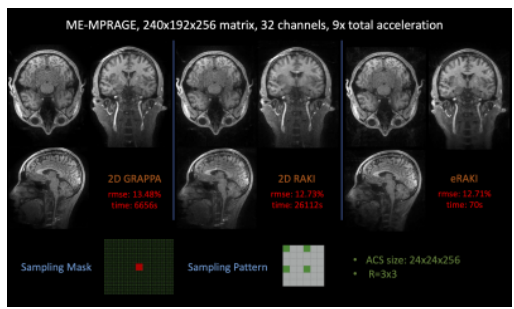

Comparisons between 2D GRAPPA, 2D RAKI, and eRAKI reconstructions on 3D ME-MPRAGE data using 24×24×256 ACS and 3×3 undersampling. We use Tikhonov regularization in GRAPPA and set parameter $\lambda=1\times e^{-9}$. Our eRAKI benefits from 3D convolutions and can achieve comparable performance with high-speed reconstruction.

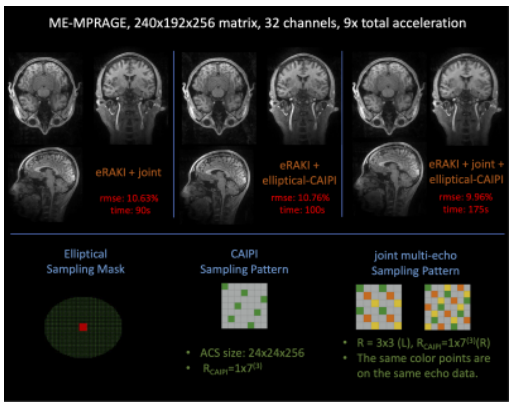

eRAKI's performance can be further improved using joint reconstruction across echoes and elliptical-CAIPI sampling. In joint reconstruction, shifted patterns are used between different echoes and all the echoes are reconstructed using one model. In this case, the number of output channels is 2×R×$N_{echo}$. Elliptical sampling provides ~4/π additional acceleration so that $R_{CAIPI}$=1×7$^{(3)}$ sampling can maintain the total acceleration of R≈9.

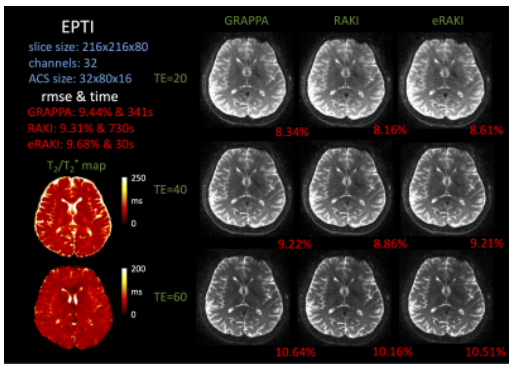

Comparisons between GRAPPA, RAKI, and eRAKI reconstructions on EPTI data using 32x80x16 ACS. All three methods are implemented in $k_x$-$k_y$-t space and eRAKI can achieve comparable performance with GRAPPA but uses only two models to perform the whole learning/reconstruction process.

| ME-MPRAGE 240x192x256 | 2D GRAPPA | 2D RAKI | eRAKI | eRAKI+ multi-echo (all 3 echoes) | ESPIRiT coil sensitivity estimation |
|---|---|---|---|---|---|
| Learning time | 40s/slice x256 slices =10240s | 100s/slice x256 slices =25600s | 30s | 75s | 50s |
| Reconstruction time | 8s/slice x256 slices =2048s | 2s/slice x256 slices =512s | 40s | 100s | |

| EPTI 216x216x80 | GRAPPA | RAKI | eRAKI | ESPIRiT coil sensitivity estimation |
|---|---|---|---|---|
| Learning time | 11s | 370s | 20s | 0.38s |
| Reconstruction time | 330s | 360s | 10s | |

Summary of learning and reconstruction time between GRAPPA, RAKI and eRAKI reconstructions. Proposed eRAKI significantly reduces the overall reconstruction time since low-resolution ESPIRiT coil sensitivities can be estimated rapidly. Note that GRAPPA is implemented on CPU while RAKI and eRAKI are implemented on GPU. But simply putting GRAPPA on gpu won't help much, since it still needs 64 different models for the typical 32 channel receiver coil data while eRAKI needs only one model.